\documentclass[runningheads]{llncs}

\usepackage[utf8]{inputenc}
\usepackage[T1]{fontenc}
\usepackage{url}
\urldef{\doilink}\url{https://doi.org/10.1007/978-3-032-39028-8_12}
\usepackage{booktabs}
\usepackage{amsfonts}
\usepackage{amssymb}
\usepackage{amsmath}
\usepackage{nicefrac}
\usepackage{graphicx}

\begin{document}

\title{Employing Discrete Fourier Transform in Representation Learning\thanks{This version of the contribution has been accepted for publication in Artificial Neural Networks in Pattern Recognition (ANNPR 2026), LNCS. The Version of Record is available online at \doilink.}}
\titlerunning{Employing DFT in Representation Learning}
\toctitle{Employing Discrete Fourier Transform in Representation Learning}

\author{Raoof HojatJalali\orcidID{0009-0004-1386-4294} \and Edmondo Trentin\orcidID{0000-0003-2197-0703}}
\authorrunning{R. HojatJalali and E. Trentin}

\institute{DIISM, Universit\`a di Siena, I-53100 Siena (Italy)\\
	\email{mohamm.hojatjalali@student.unisi.it, trentin@unisi.it}}

\maketitle

\begin{abstract}
	Image representation learning via input reconstruction is a common technique in machine learning for generating representations that can be effectively utilized by arbitrary downstream tasks. A well-established approach uses autoencoders to extract latent representations at the network's compression point. These representations are valuable because they retain essential information necessary for reconstructing the original input from the compressed latent space. In this paper, we propose an alternative learning objective. Instead of using the raw input as the reconstruction target, we employ the Discrete Fourier Transform (DFT) of the input. The DFT provides meaningful global information at each frequency level, making individual frequency components useful as separate learning targets. When dealing with multidimensional input data, the DFT offers remarkable flexibility by enabling selective transformation across specific dimensions while preserving others in the computation. Moreover, certain types of input exhibit distinct patterns in their frequency distributions, where specific frequency components consistently contain most of the magnitude, allowing us to focus on a subset of frequencies rather than the entire spectrum. These characteristics position the DFT as a viable learning objective for representation learning. We validate our approach on the CIFAR-10 and STL-10 datasets under constrained training regimes. We discovered that integrating data augmentation during pretraining enhances the learned representations. Furthermore, we demonstrate that training exclusively on the lower-frequency components, those containing the highest magnitudes, produces results comparable to using the full frequency spectrum with only minimal reductions in accuracy. Among the various configurations explored, our augmented 2D DFT method yielded the optimal performance, achieving 59.25\% and 62.56\% top-1 accuracy on CIFAR-10 and STL-10, respectively. This outperforms standard autoencoder-based reconstruction baselines, while remaining highly competitive with state-of-the-art stochastic autoencoders.

\end{abstract}

\section{Introduction}

Unsupervised representation learning continues to be a critical challenge in machine learning. Of the many methods developed in this field, three primary approaches categorized by Bengio et al. \cite{bengio2014representation} are particularly notable: probabilistic models, which infer latent variables to represent the underlying distribution of observed data; geometrically motivated manifold-learning approaches, which are based on the idea that observed data tend to cluster near a lower-dimensional manifold; and reconstruction-based algorithms such as autoencoders, which compress high-dimensional inputs into compact latent representations while preserving enough information to reconstruct the original input.

We introduce Fourier Transform Representation Learning, a method closely related to the reconstruction approaches outlined by Bengio et al. \cite{bengio2014representation}. The core idea is to use the Discrete Fourier Transform (DFT) of the input as the learning objective instead of reconstructing the original input. This approach aligns with reconstruction methods because not all such methods aim to reproduce the raw input. A notable example is Transforming Auto-encoders by Hinton et al. \cite{hinton2011transae}, where the output is an affine transformation (e.g., translation, rotation, scaling, or shearing) of the input. By treating the DFT as a mathematical transformation of the input, our approach fits naturally within the broader category of reconstruction-based learning. It is also worth emphasizing that the DFT is a lossless transformation (if the entire spectrum is used), which theoretically allows the original input data to be perfectly reconstructed by applying the inverse DFT to the output.

To rigorously evaluate this approach against existing methods, we must also consider the role of the decoder architecture. While traditional autoencoder-type models often benefit from utilizing deconvolutional (deconv) layers to reconstruct spatial inputs, our proposed DFT target formulation fundamentally relies on a Multi-Layer Perceptron (MLP) decoder to output frequency components. Therefore, in this work, we explicitly evaluate standard reconstruction baselines, including standard Autoencoders (AE), Augmentation-Augmented Autoencoders (AAAE), Variational Autoencoders (VAE), and Augmentation-Augmented Stochastic Autoencoders (AASAE), using both deconvolutional and MLP decoders to ensure a fair, comprehensive, and architecture-aware comparison against our DFT method. Additionally, we incorporate data augmentation into our DFT learning objective by default, as our experiments demonstrate it substantially improves the quality of the learned representations.

In order to contextualize the proposed approach, it is crucial to introduce first the characteristics of the input data and the chosen mathematical transformation. Section~\ref{sec:natural_images} presents the statistical properties of natural images that make them particularly suitable for our method, while Section~\ref{sec:dft_flexibility} elaborates on the flexibility of the DFT when applied to multi-dimensional spatial data.

\subsection{Natural Images as Input}
\label{sec:natural_images}

This paper centers on extracting useful representations from images as inputs. An image is constructed as a multi-dimensional numerical array, with each entry representing the intensity of a specific pixel in the image. Consider an image of size $H$-by-$W$ pixels with $C$ channels (e.g., RGB), this can be represented as an array $I$ of size $C \times H \times W$, where each element $I_{c,h,w}$ denotes the intensity of the pixel at position $(h, w)$ for channel $c$, as expressed in Equation~\eqref{eq:image_space}:

\begin{equation}
	\label{eq:image_space}
	I \in \mathbb{R}^{C \times H \times W}
\end{equation}

Particularly, we focus on the natural images, the same category of images studied in Torralba et al. \cite{torralba2003statistics}. Natural images are characterized by significant redundancy and statistical regularities. The inherent spatial redundancy in natural images makes DFT a strong candidate for learning targets, as each DFT frequency component is explicitly designed to detect a distinct periodic pattern. Another key statistical property (noted in Torralba et al.) we exploit in our experiments is that the average power spectrum of natural images reduces by $1/f^\alpha$, where $f$ is the spatial frequency. This suggests that natural images are dominated by low-frequency components, allowing training to focus on a subset of these components, as they contain most of the relevant information. This becomes crucial when working with high-resolution images, where using the full frequency spectrum could drastically increase the number of network parameters. In Section~\ref{sec:exp_selective}, we present empirical results showing that training with only lower-frequency components yields results comparable to those achieved with the complete frequency spectrum.

\subsection{Flexibility in DFT Applications}
\label{sec:dft_flexibility}

The DFT offers remarkable flexibility in its application to spatial data, enabling transformations across multiple dimensions either independently or in combination. To understand this, we begin by analyzing the general form of the n-dimensional DFT. Mathematically, it is expressed in Equation~\eqref{eq:dft_general}:

\begin{equation}
	\label{eq:dft_general}
	X_{k_1, k_2, \dots, k_n} =
	\sum_{n_1=0}^{N_1-1}
	\sum_{n_2=0}^{N_2-1} \dots
	\sum_{n_n=0}^{N_n-1}
	\left(
	x_{n_1, n_2, \dots, n_n} \cdot
	\prod_{d=1}^n e^{-2\pi i \frac{k_d n_d}{N_d}}
	\right)
\end{equation}

where $X_{k_1, k_2, ..., k_n}$ represents the frequency components, and $x_{n_1, n_2, ..., n_n}$ is the input image in the spatial domain. The indices $k_d$ and $n_d$ correspond to the frequency and spatial dimensions respectively, while $N_d$ is the size of the d-th dimension. The general form of the DFT is highly versatile, enabling transformations of data across any number of dimensions. It connects the spatial domain with the frequency domain, where periodic structures often appear more defined. The general DFT form on image space can be better understood by decomposing it hierarchically, explicitly demonstrating the connections between its various dimensional components. This becomes apparent when decomposing Equation~\eqref{eq:dft_general} into the following formulation:

\begin{equation}
	\label{eq:dft_3d}
	X_{k_c, k_h, k_w} =
	\sum_{c=0}^{C-1} \left( e^{-2\pi i \frac{k_c c}{C}}
	\sum_{h=0}^{H-1} \left( e^{-2\pi i \frac{k_h h}{H}}
	\sum_{w=0}^{W-1} \left( e^{-2\pi i \frac{k_w w}{W}} \cdot I_{c, h, w}
	\right) \right) \right)
\end{equation}

This decomposed formulation inherently utilizes the computations of lower-dimensional DFTs. It is evident that the 1D and 2D DFTs are embedded within this structure. This hierarchical structure employs a stepwise method to compute the 3D DFT. By first applying 1D DFTs along one dimension, then using these to compute 2D DFTs, and ultimately combining them for the full 3D DFT result. As we transition from 1D to 2D and finally to 3D, we gradually integrate more dimensions of the image space Equation~\eqref{eq:image_space}. This hierarchical multi-dimensional DFT method facilitates the generation of tailored learning targets for our experiments. When applying a 1D DFT to rows, the computation is limited to this dimension, which helps detect horizontal frequency patterns while keeping spatial information intact in columns and channels. By scaling this to a 2D DFT across the image's plane, computation is restricted to both rows and columns, ensuring that channel-specific information remains unaffected. Finally, a 3D DFT incorporates all dimensions, where information is shared across rows, columns, and channels.

\section{Related Works}
\label{sec:related_works}

\subsection{Frequency Spectrum and Model Generalization}

The frequency spectrum of input data provides critical insights into the generalization behaviors and feature preferences of Convolutional Neural Networks (CNNs). In \textit{High-Frequency Component Helps Explain the Generalization of Convolutional Neural Networks}, Wang et al. (2020) investigate how CNNs process different frequency components, demonstrating that these models can exploit high-frequency patterns that are ``almost imperceptible to a human'' \cite{Wangetal2020}. This perceptional disparity explains unintuitive model behaviors, such as adversarial vulnerability, where predictions are ``almost entirely determined by the high-frequency components of the image'' \cite{Wangetal2020}. To evaluate the viability of different frequency bands, they trained models exclusively on high- and low-frequency counterparts of images, finding that the low-frequency component is ``much more `generalizable' than HFC'' and aligns with human visual preference \cite{Wangetal2020}. Their analysis suggests that leveraging low-frequency signals not only leads to ``the steepest descent of the loss surface'' but also provides a more robust alignment between model perception and human annotation \cite{Wangetal2020}.

\subsection{Neural Networks and Frequency Domain Inputs}

The application of frequency domain transformations to neural network architectures provides a mechanism to mitigate the information loss associated with spatial downsampling. In \textit{Learning in the Frequency Domain}, Xu et al. (2020) propose feeding Discrete Cosine Transform (DCT) coefficients directly into CNNs rather than using traditional spatial RGB inputs \cite{9157408}. Through a learning-based channel selection method, they identify a ``spectral bias'' within neural networks, observing that CNN models ``are more sensitive to low-frequency channels than high-frequency channels,'' which aligns with the characteristics of the human visual system \cite{9157408}. By leveraging this statistical property, their framework statically selects these lower-frequency components to discard ``trivial frequency components'' prior to inference, reducing input data size while preserving the essential patterns necessary for downstream image understanding tasks \cite{9157408}.

\subsection{Dimensionality Reduction using Deep Autoencoders}

The extraction of low-dimensional representations through input reconstruction is a foundational concept in neural network architectures. In \textit{Reducing the Dimensionality of Data with Neural Networks}, Hinton and Salakhutdinov (2006) demonstrate that ``high-dimensional data can be converted to low-dimensional codes by training a multilayer neural network with a small central layer to reconstruct high-dimensional input vectors'' \cite{doi:10.1126/science.1127647}. They identify that optimizing deep autoencoders is inherently difficult; depending on the initialization, models typically ``find poor local minima'' or suffer from ``tiny'' gradients in early layers, making gradient descent infeasible \cite{doi:10.1126/science.1127647}. To address this, they introduce a ``pretraining'' procedure using restricted Boltzmann machines (RBMs) to learn ``one layer of features at a time'' \cite{doi:10.1126/science.1127647}. Following this initialization, the pre-trained model is ``unfolded'' to produce symmetric ``encoder and decoder networks'' \cite{doi:10.1126/science.1127647}. This structured initialization strategy provides weights close to a good solution, allowing for global fine-tuning via backpropagation through the entire autoencoder for optimal reconstruction, yielding representations that outperform standard linear methods like principal components analysis \cite{doi:10.1126/science.1127647}.

\section{Fourier Transform Representation Learning}

\begin{figure}
	\centering
	\includegraphics[width=0.8\textwidth]{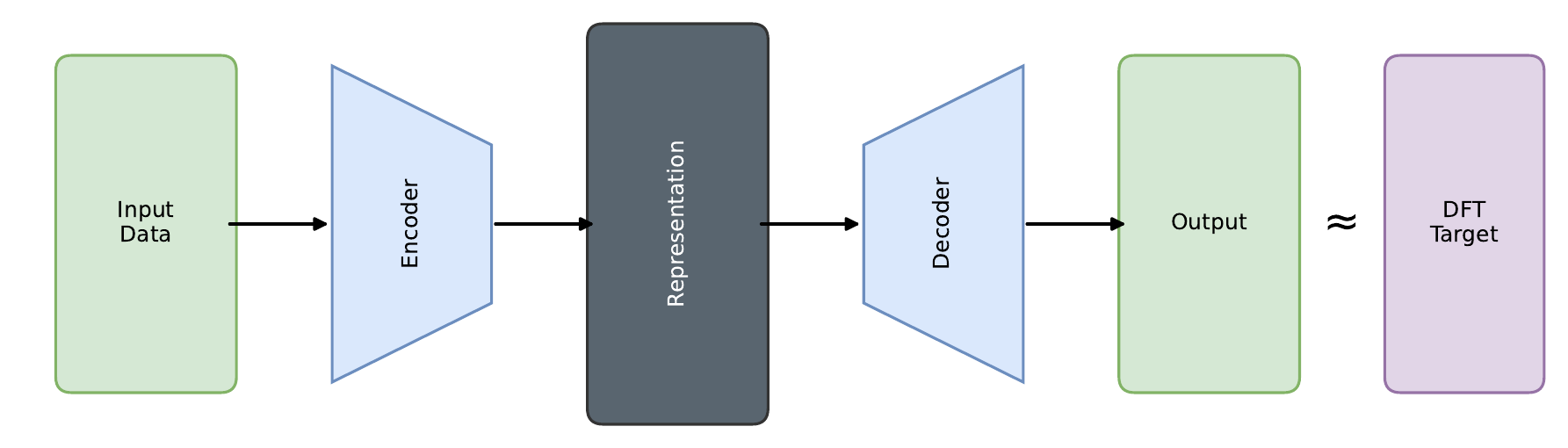}
	\caption{DFT representation learning employs a similar autoencoder architecture to reconstruct frequency components derived from the DFT.}
	\label{fig:normal_arch}
\end{figure}

Having explored the mathematical foundations of the DFT and its application to our image space, we now introduce our proposed approach for learning image representation. As shown in Figure~\ref{fig:normal_arch}, the core idea involves using an autoencoder to compress the input into a latent representation. From this compressed form, we reconstruct the DFT frequency components that originated from the input images.

\subsection{Learning Objectives}
\label{sec:learning_obj}

The DFT produces a complex-valued output. As our network outputs real values, we must identify a way to express DFT frequencies in a real-valued form. In addition to real and imaginary parts, frequency analysis typically uses magnitude and phase to describe complex values. If we denote the DFT output as $X[k]$ (where $k$ represents the frequency index across any dimension), we can express the relationship between these components using the decompositions expressed in Equation~\eqref{eq:dft_components}:

\begin{equation}
	\label{eq:dft_components}
	X[k]
	= \text{Re}(X[k]) + i \cdot \text{Im}(X[k])
	= |X[k]| \cdot e^{i \cdot \angle X[k]}
\end{equation}

Each component isolates a unique aspect of the frequency domain, highlighting different characteristics. The real and imaginary parts, when combined, fully define the complex DFT output, while individually they reflect projections onto their respective axes. The magnitude measures the distance from the origin to the complex point in the plane, signifying the strength, while the phase is the angle from the real axis, reflecting the shift in the periodic pattern captured by a DFT frequency component. To evaluate which components yield more effective learning representations, we defined 6 learning targets:

\begin{itemize}
	\item \textbf{Real Part Only}: Uses only the real part of the frequency domain $\text{Re}(X[k])$ as the target, isolating real-value contributions to learning.

	\item \textbf{Imaginary Part Only}: Uses only the imaginary part of the frequency domain $\text{Im}(X[k])$ as the target, isolating imaginary-value contributions to learning.

	\item \textbf{Real and Imaginary}: Using concatenation of both real and imaginary parts as targets.

	\item \textbf{Magnitude Only}: Uses only the magnitude of frequency components $|X[k]|$ as the target, focusing on strength while ignoring phase.

	\item \textbf{Phase Only}: Uses only the phase information of frequency components $\angle X[k]$ as the target, focusing on shift of the periodic patterns.

	\item \textbf{Magnitude and Phase}: Using concatenation of both magnitude and phase as targets.
\end{itemize}

\subsection{DFT Dimensionality Selection}
\label{sec:dft_dim}

Building upon the DFT formulation adapted to image space \ref{eq:dft_3d}, we explored three distinct learning objectives: a 1D DFT applied along the row dimension, capturing horizontal frequency patterns while preserving spatial information in column and channel dimensions; a 2D DFT covering both row and column dimensions, while maintaining cross-channel information; and a complete 3D DFT encompassing all three dimensions of image space. Each approach represents a different level of frequency information extraction from the input data, allowing us to evaluate the effectiveness of representation learning based on the dimensionality selection of the DFT.

\subsection{Specific Frequency Components of Interest}
\label{sec:specific_freq}

\begin{figure}
	\centering
	\includegraphics[width=0.4\textwidth]{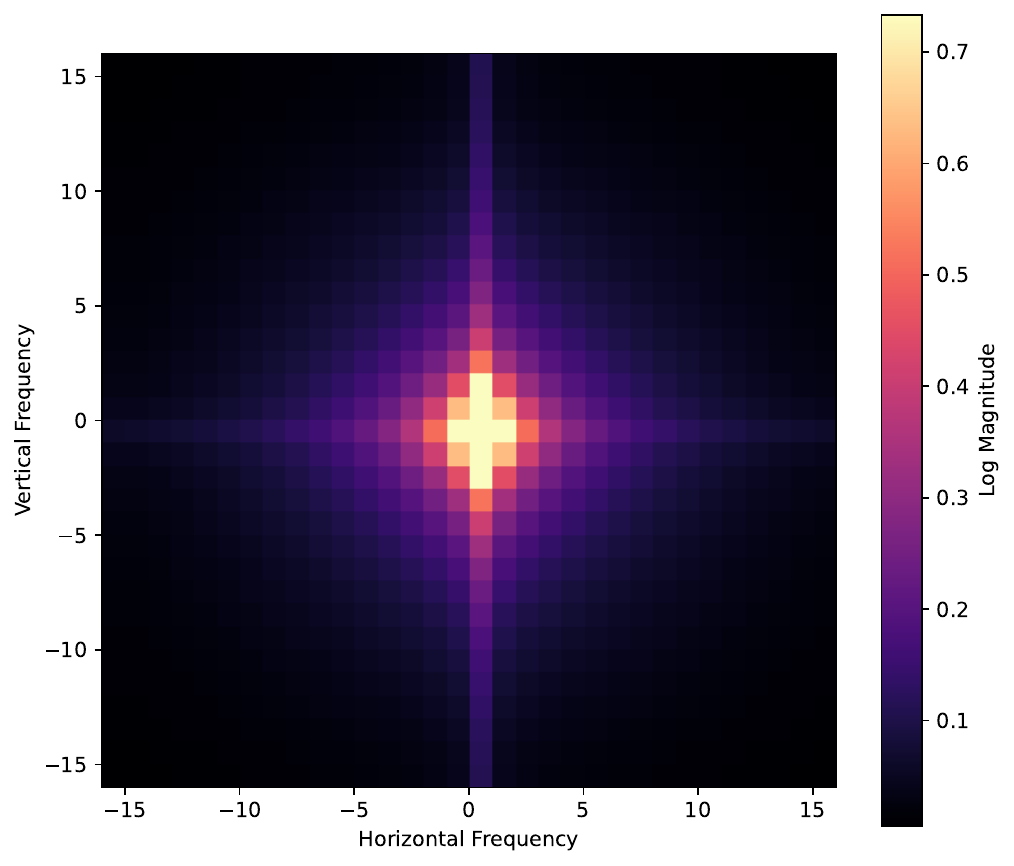}
	\caption{Aggregated 2D Magnitude Spectrum for the CIFAR-10 dataset.}
	\label{fig:dft_2d_magnitude_aggr_dist}
\end{figure}

As mentioned in Section~\ref{sec:natural_images}, natural images are characterized by their dominance of low-frequency components in their magnitude distribution. To demonstrate this, we performed a 2D DFT across spatial dimensions for all images in the CIFAR-10 dataset \cite{cifar10}, as detailed in Equation~\eqref{eq:dft_aggregation}:

\begin{equation}
	\label{eq:dft_aggregation}
	A_{k_h, k_w} =
	\frac{1}{N \times C}
	\sum_{n=0}^{N-1}
	\sum_{c=0}^{C-1} |X^{(n)}_{c,k_h,k_w}|
\end{equation}

where $A_{k_h, k_w}$ represents the aggregated magnitude for frequency components $(k_h, k_w)$, $N$ is the dataset size, and $X^{(n)}_{c,k_h,k_w}$ is the 2D DFT component for channel $c$ of image $n$. Furthermore, because image inputs are strictly real-valued, their 2D DFT exhibits Hermitian (conjugate) symmetry. Consequently, the magnitude spectrum is symmetric (even), meaning $|X_{k_h, k_w}| = |X_{-k_h, -k_w}|$. Since half the spectrum is redundant, we safely discard it. Figure~\ref{fig:dft_2d_magnitude_aggr_dist} plots the unique components (shifted to center zero-frequency), revealing peak magnitudes densely clustered at the low-frequency center. We leveraged this in Section~\ref{sec:exp_selective} by training exclusively on the lowest quarter and eighth of the spectrum (extracted via a square crop around the DC component), proving these subsets provide highly effective representations.

\section{Experiments}

In this section, we present experiments conducted to evaluate the effectiveness of the Fourier Transform Representation Learning approach. Each experiment investigates a specific aspect of the method, revealing how the learned representation is shaped and utilized for downstream tasks.

To evaluate the learned representations, we applied the linear evaluation protocol adapted from the AASAE methodology \cite{falcon2021aasae}. This protocol freezes the encoder's parameters and trains a single linear classifier on top of the frozen representation. Due to hardware limitations, all of our pretraining experiments are constrained to specific epoch limits depending on the dataset. A detailed overview of the implementation settings, including the network architectures and hyper-parameter configurations utilized in the subsequent experiments, is provided in Section~\ref{sec:impl_details}.

To systematically isolate and evaluate the core components of our proposed approach, our experiments are structured in two phases. In the first phase, we conduct extensive ablation studies (Sections 4.1--4.3) exclusively on the CIFAR-10 dataset using a strictly controlled, unaugmented training regime (both clean inputs and clean targets). This ensures that the observed performance differences are strictly attributable to the DFT target formulations and dimensionalities themselves, rather than complex interactions with data augmentation. In the second phase, after empirically establishing the optimal base configuration, we evaluate the impact of data augmentation (Section 4.4) and compare our optimized model against standard baselines on both CIFAR-10 \cite{cifar10} and STL-10 \cite{coates2011analysis} (Section 4.5).

\subsection{Learning Objectives Experiments}

In this experiment, we examine different target formulations derived from the 2D DFT across both row and column dimensions. As highlighted in Section~\ref{sec:learning_obj}, the goal is to identify which DFT-derived target generates the most effective learned representations. As shown in Table \ref{tab:dft_target_comparison}, across six different target formulations, the Magnitude representation is the most effective. The Magnitude-only target achieved the highest downstream accuracy of 56.04\% on CIFAR-10, outperforming all other formulations. Therefore, in all subsequent experiments, we exclusively use the Magnitude representation as our DFT target, as it provides optimal performance.

\begin{table}
	\centering
	\caption{Performance Comparison of Different DFT Target Formulations on CIFAR-10.}
	\label{tab:dft_target_comparison}
	\setlength{\tabcolsep}{10pt}
	\begin{tabular}{|l|c|}
		\hline
		DFT Target        & Accuracy (\%)  \\
		\hline
		Real Part         & 41.30          \\
		Imaginary Part    & 37.93          \\
		Real + Imaginary  & 42.85          \\
		Magnitude         & \textbf{56.04} \\
		Phase             & 45.58          \\
		Magnitude + Phase & 51.23          \\
		\hline
	\end{tabular}
\end{table}

\subsection{DFT Dimensionality Selection Experiments}

Next, we systematically examined how varying the dimensionality of the DFT impacts the quality of learned representations. Building on the theoretical foundations outlined in Section~\ref{sec:dft_dim}, we analyzed three distinct approaches: a 1D DFT applied along the row dimension, a 2D DFT incorporating both row and column dimensions, and a full 3D DFT encompassing all three dimensions. As shown in Table \ref{tab:dft_dim_comparison}, the 2D DFT configuration achieves superior performance compared to the 1D and 3D approaches, yielding the highest accuracy of 56.04\% on CIFAR-10. Notably, the 3D DFT, despite incorporating additional dimensional information, performs worse than both the 1D and 2D methods. We attribute this suboptimal performance to the inherent differences between spatial and channel dimensions. While rows and columns exhibit strong spatial correlations and structural patterns, which are features that frequency transformations can effectively leverage, the channel dimension lacks these intrinsic spatial relationships.

\begin{table}
	\centering
	\caption{Performance Comparison of Different DFT Dimensionalities on CIFAR-10.}
	\label{tab:dft_dim_comparison}
	\setlength{\tabcolsep}{10pt}
	\begin{tabular}{|l|c|}
		\hline
		DFT Dim & Accuracy (\%)  \\
		\hline
		1D      & 50.53          \\
		2D      & \textbf{56.04} \\
		3D      & 55.63          \\
		\hline
	\end{tabular}
\end{table}

\subsection{Selective Frequency Components Experiments}
\label{sec:exp_selective}

Building on Section~\ref{sec:specific_freq}, we evaluated selective frequency targets. Exploiting Hermitian symmetry and low-frequency dominance, we trained models using only the lowest quarter and eighth of the 2D DFT spectrum (applied as square crops around the DC component). Table~\ref{tab:selective_dft_comparison} shows the full 2D DFT yields the highest accuracy (56.04\%), but the cropped subsets exhibit only modest declines (54.00\% and 53.51\%). Despite isolating strictly the lowest frequencies, these highly compressed targets achieve competitive performance against the full spectrum baseline.

\begin{table}
	\centering
	\caption{Performance Comparison of Frequency Selectivity on CIFAR-10.}
	\label{tab:selective_dft_comparison}
	\setlength{\tabcolsep}{10pt}
	\begin{tabular}{|l|c|}
		\hline
		DFT Target Selection & Accuracy (\%)  \\
		\hline
		2D Full Spectrum     & \textbf{56.04} \\
		2D Quarter Spectrum  & 54.00          \\
		2D Eighth Spectrum   & 53.51          \\
		\hline
	\end{tabular}
\end{table}

\subsection{Impact of Data Augmentation}
\label{sec:exp_augmentation}

Having empirically established the optimal configuration (2D DFT with Magnitude targets) on clean inputs, we evaluated the impact of integrating data augmentation into the training pipeline. Specifically, we apply the exact same augmentation strategy employed by the AAAE and AASAE baselines to the input images. We observed a notable performance increase using this configuration. On the CIFAR-10 dataset using our 2D Magnitude MLP architecture, the model trained on clean inputs (as reported in Table~\ref{tab:dft_dim_comparison}) achieved 56.04\% accuracy. When trained with this augmentation strategy, where the input is augmented but the target remains clean, the accuracy improved to 59.25\%. Given this substantial gain, we adopted this augmented DFT approach as our default configuration for all subsequent comparative baseline experiments to ensure a rigorous and fair evaluation.

\subsection{Comparative Representation Learning Experiments}

Finally, we compare our Fourier Transform Representation Learning approach against several standard reconstruction-based baselines: the traditional AE, AAAE, VAE, and AASAE \cite{falcon2021aasae}. Standard autoencoder-type models often benefit from using deconvolutional (deconv) layers for spatial reconstruction. To ensure a fair comparison, we evaluate these baselines using both deconv and MLP decoders, contrasting them directly with our DFT approach, which strictly utilizes an MLP decoder.

As shown in Tables \ref{tab:representation_comparison_cifar} and \ref{tab:representation_comparison_stl}, our augmented 2D DFT method achieves 59.25\% accuracy on CIFAR-10 and 62.56\% on STL-10. This performance outperforms the standard AE, AAAE, and VAE baselines on both datasets. Notably, on CIFAR-10, our method outperforms these baselines regardless of whether they employ deconvolutional or MLP decoders. Most notably, our approach outperforms the MLP-based AASAE (which achieves 56.60\% on CIFAR-10), demonstrating that when constrained to identical decoder architectures, the DFT target yields superior representations. While our method performs slightly below the deconvolutional AASAE baseline under these constrained training regimes, the results clearly establish the DFT as a highly competitive and flexible alternative for representation learning.

\begin{table}
	\centering
	\caption{Performance Comparison of Representation Learning Methods on CIFAR-10.}
	\label{tab:representation_comparison_cifar}
	\setlength{\tabcolsep}{10pt}
	\begin{tabular}{|l|l|c|}
		\hline
		Method        & Decoder & Accuracy (\%)  \\
		\hline
		AE            & Deconv  & 43.79          \\
		AE            & MLP     & 41.78          \\
		AAAE          & Deconv  & 40.03          \\
		AAAE          & MLP     & 39.29          \\
		VAE           & Deconv  & 42.70          \\
		VAE           & MLP     & 44.49          \\
		AASAE         & Deconv  & \textbf{62.83} \\
		AASAE         & MLP     & 56.60          \\
		\hline
		2D DFT (Ours) & MLP     & \textbf{59.25} \\
		\hline
	\end{tabular}
\end{table}

\begin{table}
	\centering
	\caption{Performance Comparison of Representation Learning Methods on STL-10.}
	\label{tab:representation_comparison_stl}
	\setlength{\tabcolsep}{10pt}
	\begin{tabular}{|l|l|c|}
		\hline
		Method        & Decoder & Accuracy (\%)  \\
		\hline
		AE            & Deconv  & 36.78          \\
		AAAE          & Deconv  & 36.42          \\
		VAE           & Deconv  & 35.82          \\
		AASAE         & Deconv  & \textbf{65.87} \\
		\hline
		2D DFT (Ours) & MLP     & \textbf{62.56} \\
		\hline
	\end{tabular}
\end{table}

\subsection{Implementation Details}
\label{sec:impl_details}

In this section, we outline the implementation details for our experiments. The datasets used are CIFAR-10 \cite{cifar10} and STL-10 \cite{coates2011analysis}, with labels removed during the pretraining phase. We adopt an architectural setup based on the AASAE framework, featuring a ResNet-50 \cite{he2016resnet} encoder with 2048 output units. For the baseline models (AE, AAAE, VAE, AASAE), we experiment with both standard deconvolutional decoders and MLP decoders. For our DFT method, the decoder strictly comprises a Multi-Layer Perceptron (MLP) designed to produce outputs with the same dimensionality as the DFT targets. For the loss function in our approach, we compute the squared differences between the model's outputs and the DFT targets. Due to hardware limitations, all models across all experiments are pretrained under constrained regimes. Specifically, for CIFAR-10, models are trained for 90 epochs with a batch size of 256 and a learning rate of $2.5 \times 10^{-4}$. For STL-10, models are trained for 60 epochs with a batch size of 256 and a learning rate of $5 \times 10^{-4}$. Furthermore, based on empirical performance, the same data augmentation applied to AASAE is applied by default across all applicable DFT and baseline configurations to ensure robust representation learning.

\section{Conclusion}

In this paper, we proposed Fourier Transform Representation Learning, a novel reconstruction-based approach to unsupervised representation learning. We discovered that among various DFT target formulations, the magnitude representation consistently emerged as the superior learning target. Furthermore, our investigation into DFT dimensionalities revealed that a 2D formulation delivered optimal performance. Leveraging this, we explored selective frequency component training. Exploiting Hermitian symmetry and low-frequency magnitude dominance, we discarded redundant components and restricted targets to the lowest quarter or eighth of the spectrum. While the complete spectrum yielded the highest accuracy, these heavily compressed subsets achieved remarkably competitive results with minimal performance drops, confirming the dense informational value of low frequencies.

A pivotal finding of this work is the impact of simulated data augmentation on the frequency domain targets. By applying standard augmentations to the input images while computing the DFT targets from the \textit{clean} originals, the model is forced to learn robust, invariant representations. This configuration yielded substantial performance gains and was adopted as our primary evaluation method.

Finally, when evaluating this augmented 2D DFT approach against established reconstruction-based methods on CIFAR-10 and STL-10, our model outperformed standard baselines (AE, AAAE, and VAE), regardless of whether they employed deconvolutional or MLP decoders. Most crucially, when comparing models constrained to identical MLP architectures, our DFT approach outperformed the AASAE baseline. While the full deconvolutional AASAE model retains a slight performance edge under our constrained training regimes, our results conclusively establish the Discrete Fourier Transform as a highly competitive, flexible, and robust learning objective for representation learning.

\begin{credits}
	\subsubsection{\discintname}
	The authors have no competing interests to declare that are relevant to the content of this article.
\end{credits}

\bibliographystyle{splncs04}
\bibliography{references}

\end{document}